\documentclass[runningheads]{DASFAA}
\usepackage[T1]{fontenc}
\usepackage{hyperref}
\usepackage{url}
\usepackage{graphicx}
\usepackage{booktabs}
\usepackage{multirow}
\usepackage{colortbl}
\usepackage{xcolor}
\usepackage{array}
\usepackage{amsmath}  
\usepackage{multicol} 
\usepackage{threeparttable} 
\usepackage{float}
\usepackage{wrapfig}
\usepackage{cite}
\usepackage{subcaption}
\usepackage{amssymb}
\usepackage[misc]{ifsym}
\usepackage{nopageno}
\usepackage{verbatim}
\begin{document}
\title{Enhancing Time Series Forecasting via Multi-Level Text Alignment with LLMs}
%
%
\author{Taibiao Zhao, Xiaobing Chen, and Mingxuan Sun\textsuperscript{\Letter}}
\institute{Louisiana State University, Baton Rouge, LA 70803, USA
\email{\{tzhao3,xchen87,msun11\}@lsu.edu}\\}
\maketitle              
\begin{abstract}
The adaptation of large language models (LLMs) to time series forecasting poses unique challenges, as time series data is continuous in nature, while LLMs operate on discrete tokens. Despite the success of LLMs in natural language processing (NLP) and other structured domains, aligning time series data with language-based representations while maintaining both predictive accuracy and interpretability remains a significant hurdle. Existing methods have attempted to reprogram time series data into text-based forms, but these often fall short in delivering meaningful, interpretable results.
In this paper, we propose a multi-level text alignment framework for time series forecasting using LLMs that not only improves prediction accuracy but also enhances the interpretability of time series representations. Our method decomposes time series into trend, seasonal, and residual components, which are then reprogrammed into component-specific text representations. 
We introduce a multi-level alignment mechanism, where component-specific embeddings are aligned with pre-trained word tokens, enabling more interpretable forecasts. 
Experiments on multiple datasets demonstrate that our method outperforms state-of-the-art models in accuracy while providing good interpretability. Our code is available at \url{https://github.com/ztb-35/MLTA}.

\end{abstract}

\section{Introduction}


Large language models (LLMs), trained on vast and diverse text corpora, provide a powerful foundation for various downstream tasks, requiring only minimal task-specific prompt engineering or fine-tuning. This flexibility has sparked a growing interest in leveraging LLMs for time series analysis. For example, methods like Promptcast~\cite{xue2023promptcast} and LLMTime~\cite{gruver2024LLMTime} reformulate numerical inputs and outputs into prompts, treating time series forecasting as a sentence-to-sentence task, which enables the direct application of LLMs. Meanwhile, approaches like TEMPO~\cite{caotempo} and GPT4TS~\cite{gpt4ts} take a different route by fine-tuning pre-trained LLMs, modifying components such as the Add\&Norm layers and positional embeddings, further demonstrating LLMs’ adaptability for time series forecasting.

\begin{figure}[t]
    \centering
    \begin{minipage}{1\textwidth}
    \hspace*{0pt}
    \begin{subfigure}[t]{0.2\textwidth} 
        \centering
        \includegraphics[height=1in]{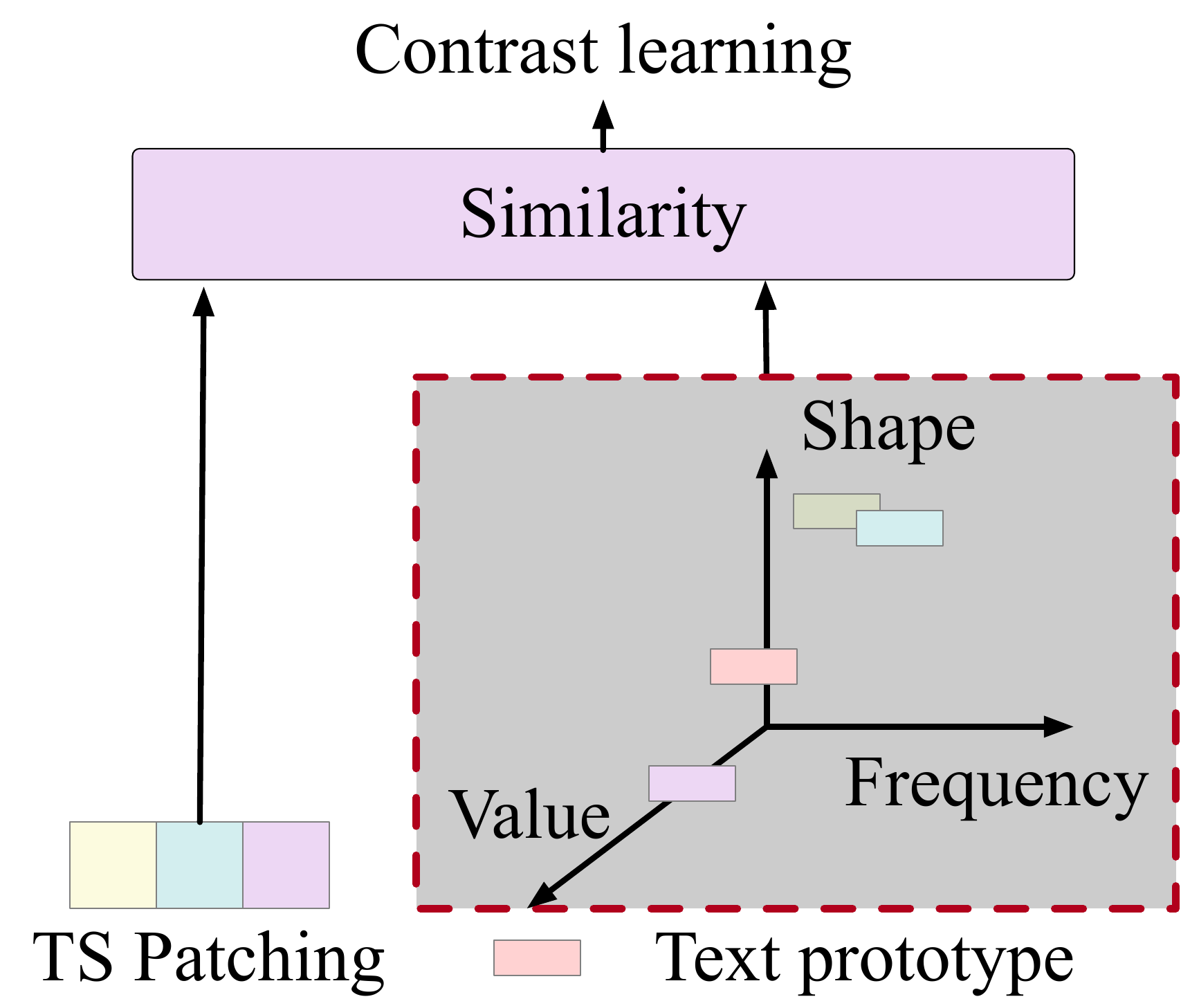} 
        \caption{}
        \label{fig:subfig1}
    \end{subfigure}
    \hspace{0.03\textwidth} 
    \begin{subfigure}[t]{0.2\textwidth} 
        \centering
        \includegraphics[height=1in]{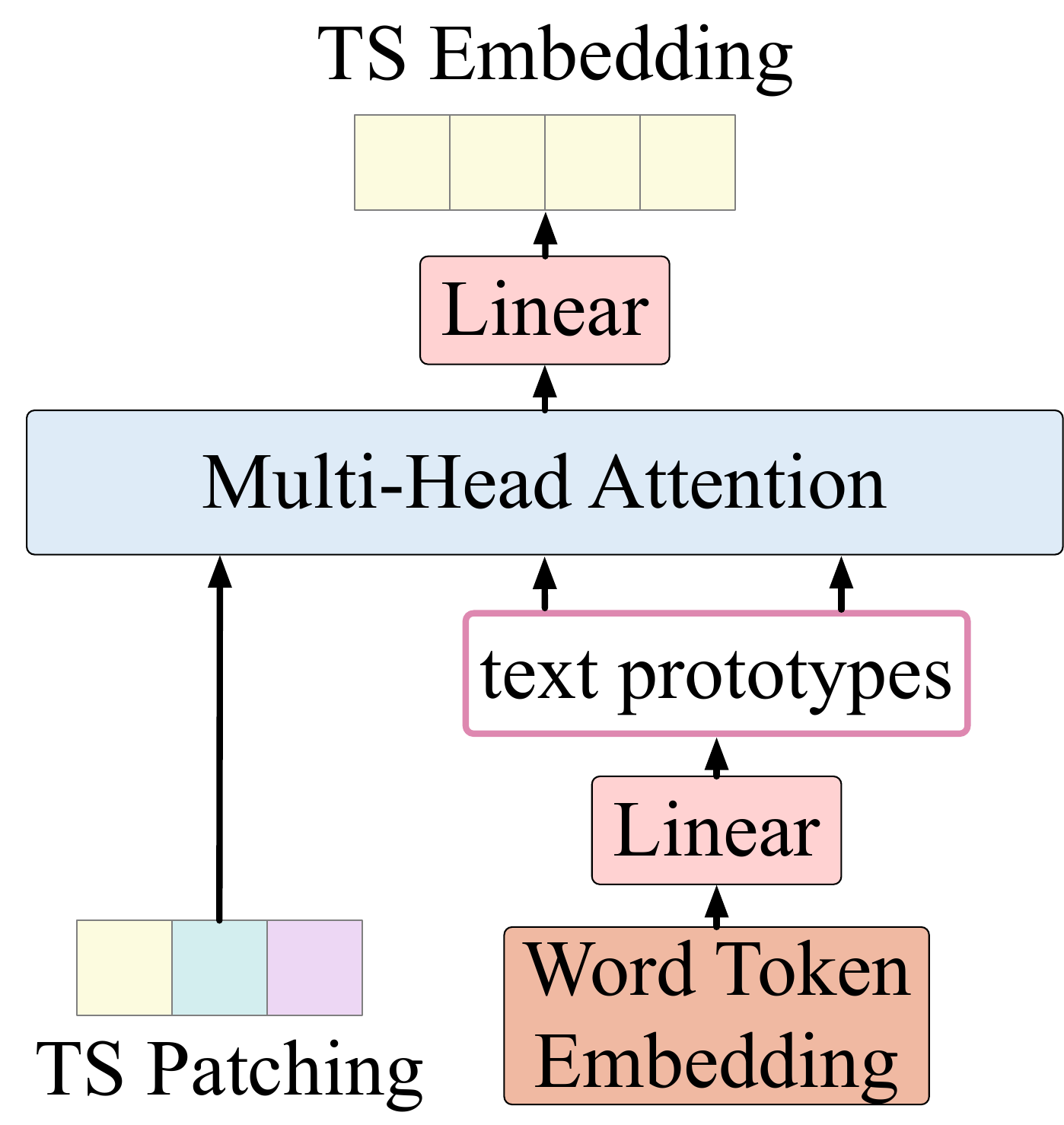} 
        \caption{}
        \label{fig:subfig2}
    \end{subfigure}
    \hspace{0.0\textwidth} 
    \begin{subfigure}[t]{0.2\textwidth} 
        \centering
        \includegraphics[height=1in]{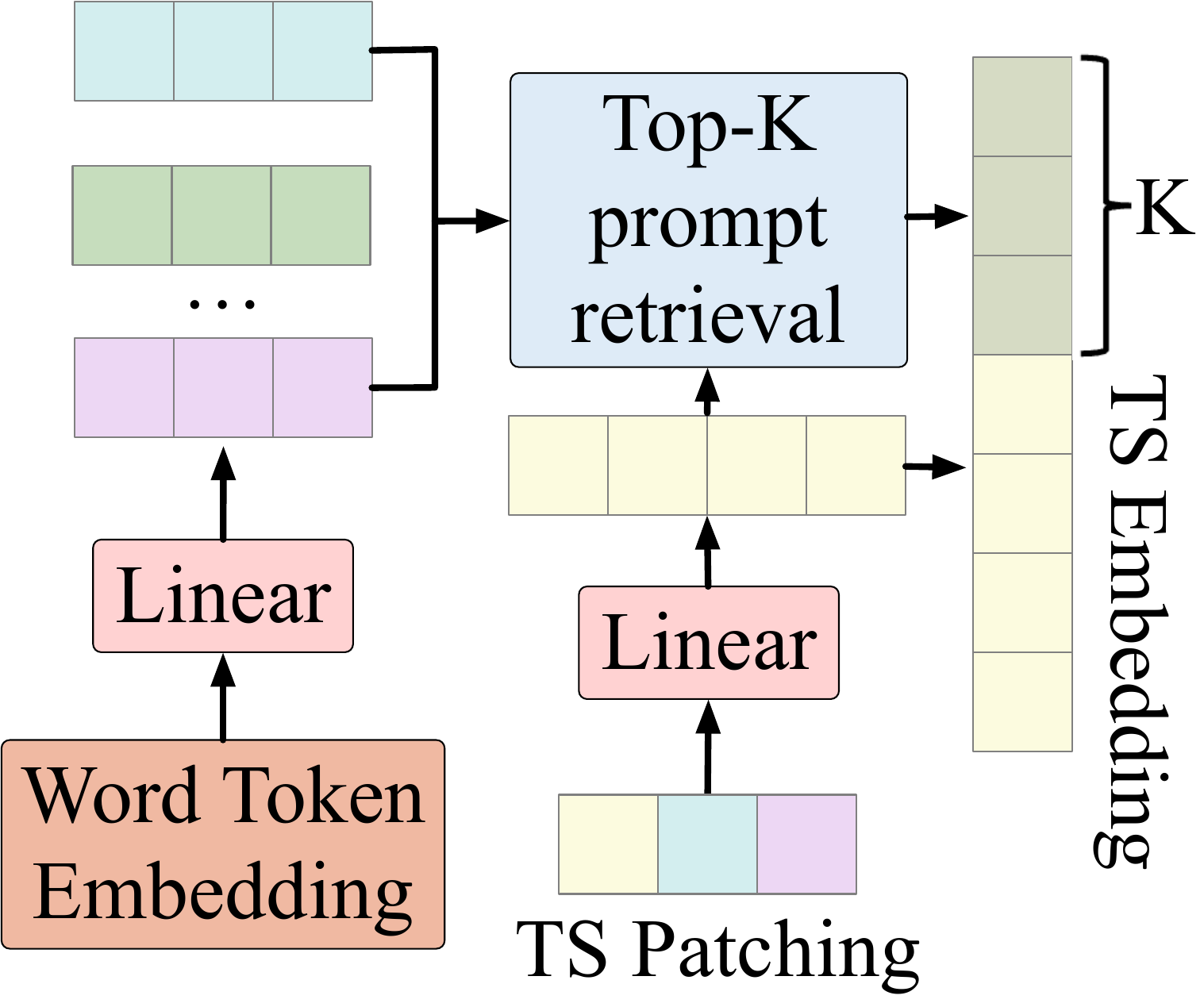} 
        \caption{}
        \label{fig:subfig3}
    \end{subfigure}
    \hspace{0.03\textwidth}
    \begin{subfigure}[t]{0.2\textwidth} 
        \centering
        \includegraphics[height=1in]{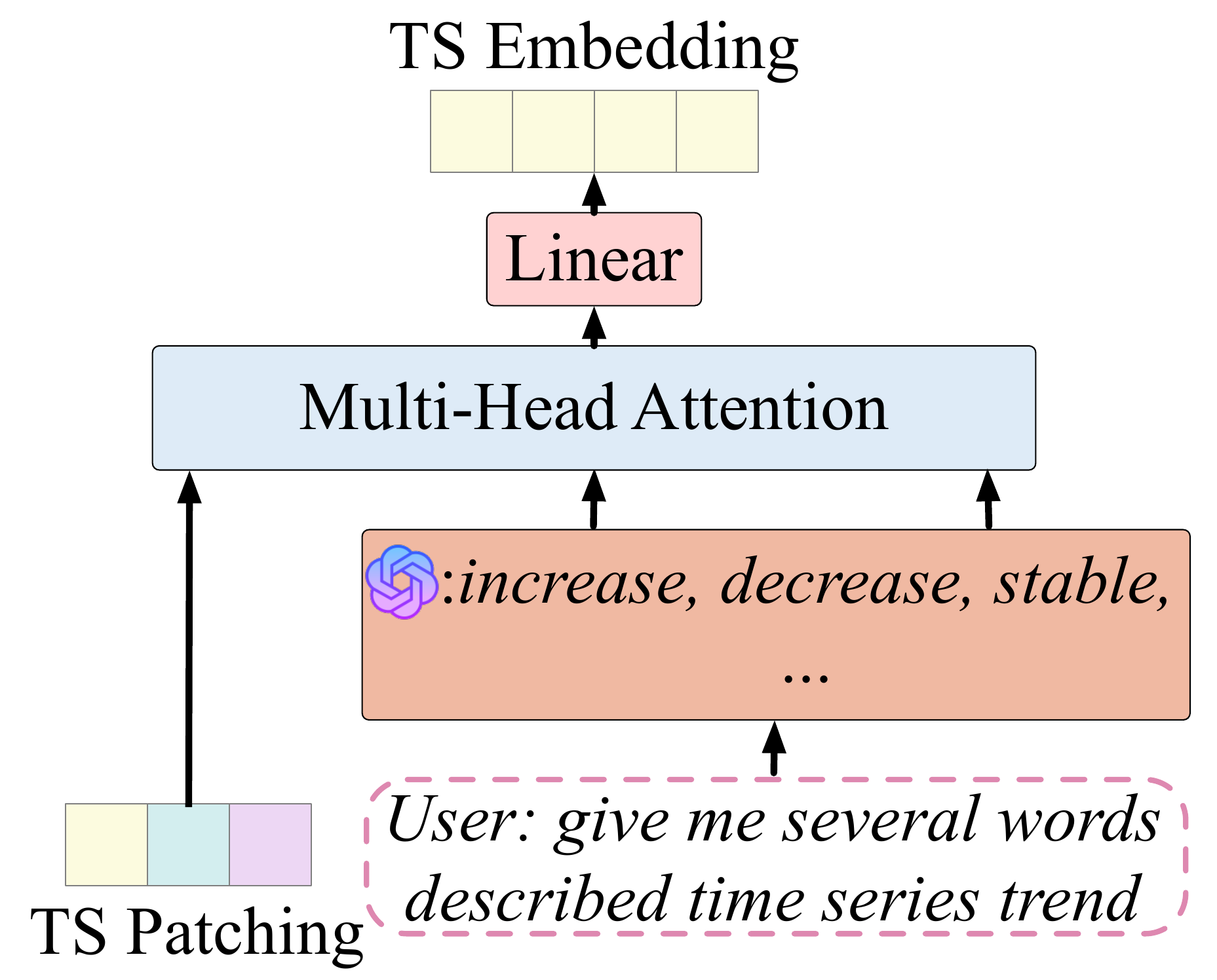} 
        \caption{}
        \label{fig:subfig4}
    \end{subfigure}
\end{minipage}
    \caption{Cross-modality time series embeddings of \((a)\) contrast learning of text-prototype-aligned time series embeddings, \((b)\) text prototypes reprogramming,  \((c)\)semantic informed prompt and and \((d)\) proposed anchors alignment of our multi-level alignment.}
\label{cross_modalities}
\vspace{-1em}
\end{figure}

Despite their potential, the benefits of LLMs in time series forecasting depend on the effective alignment between time series data and natural language modalities. TEST~\cite{sun2023test} maps time series to word embeddings via contrastive alignment (Figure \ref{fig:subfig1}), while TimeLLM~\cite{timellm} reprograms patches into text prototypes (Figure \ref{fig:subfig2}). S$^{2}$IP-LLM~\cite{s2ip} (Figure \ref{fig:subfig3}) employs semantic prompts to bridge embeddings and text. These approaches, however, primarily achieve a ``time series$\rightarrow$pattern$\rightarrow$text'' transformation to activate LLMs for time series tasks. This process often leads to unexpected outcomes. For example, the embedding of a subsequence with an upward trend may be misaligned with a word representing a decline or with a word that doesn't capture the trend at all. As a result, the challenge remains to fully unlock LLMs’ capabilities for general time series forecasting in a way that is both accurate and interpretable.

In this paper, we address the challenge of interpretability in LLM-based time series forecasting by developing an interpretable multi-level text alignment framework while preserving the backbone model. Our approach consists of two key principles: (a) modeling time series components—trend, seasonality, and residuals, and (b) deriving interpretable explanations via multi-level text alignment. Using additive decomposition methods, like LOESS~\cite{cleveland1990stl}, we decompose time series into three additive components, reprogramming them into component-specific anchors for better LLM alignment (Figure \ref{fig:subfig4}). Additionally, we employ component-specific prompts to guide the generation of learnable continuous vector representations that encode temporal knowledge of each component.

In summary, the main contributions of this paper are as follows: (1) We propose an interpretable multi-level text alignment framework for time series forecasting using LLMs while keeping the backbone model unchanged. (2) Our method leverages this multi-level alignment to map decomposed time series components—trend, seasonality, and residuals—into distinctive, informative joint representations. The aligned trend-specific anchors enhance the interpretability of LLMs, while the aligned seasonality and residual prototypes improve the overall representation of the input time series. (3) Experimental results on multiple datasets validate the superiority of our model over state-of-the-art approaches, highlighting the effectiveness of interpretable multi-level text alignment.

\section{Related work}

Recent advancements in LLMs like GPT-2~\cite{gpt2} and LLaMA~\cite{touvron2023llama} have enabled new possibilities for time series modeling. Efforts to adapt LLMs for this domain include converting time series into text~\cite{xue2023promptcast}, tokenizing data into patches for fine-tuning~\cite{gpt4ts,gruver2024LLMTime}, and leveraging retrieval-based prompts~\cite{caotempo,s2ip}. However, these approaches often lack interpretability, treating time series as mere token sequences and ignoring temporal structures. This misalignment can lead to inaccurate predictions and reduced model transparency, particularly for multivariate time series. To address this, we propose an interpretable multi-level text alignment framework that preserves the integrity of pre-trained LLMs while effectively aligning time series components with anchors.

A key challenge in adapting LLMs for time series forecasting is aligning continuous data with discrete token embeddings. Inspired by prototype-level contrast methods, works like~\cite{sun2023test} select text embeddings as prototypes (Figure \ref{fig:subfig1}), while~\cite{timellm} reprogram time series using source modality prompts without altering the backbone LLM (Figure \ref{fig:subfig2}). These methods follow a "time series → pattern → text" paradigm, but their arbitrary prototype selection may not capture the true characteristics of the data~\cite{sun2023test}. More detailed related works see full paper~\cite{FullPaper}. To overcome this, our approach uses time series-specific anchors (e.g., "increase, decrease, stable") to guide token embedding learning (Figure \ref{fig:subfig4}), enhancing both interpretability and forecasting performance. 

\begin{figure*}[t]
\centering
\includegraphics[width=\linewidth]{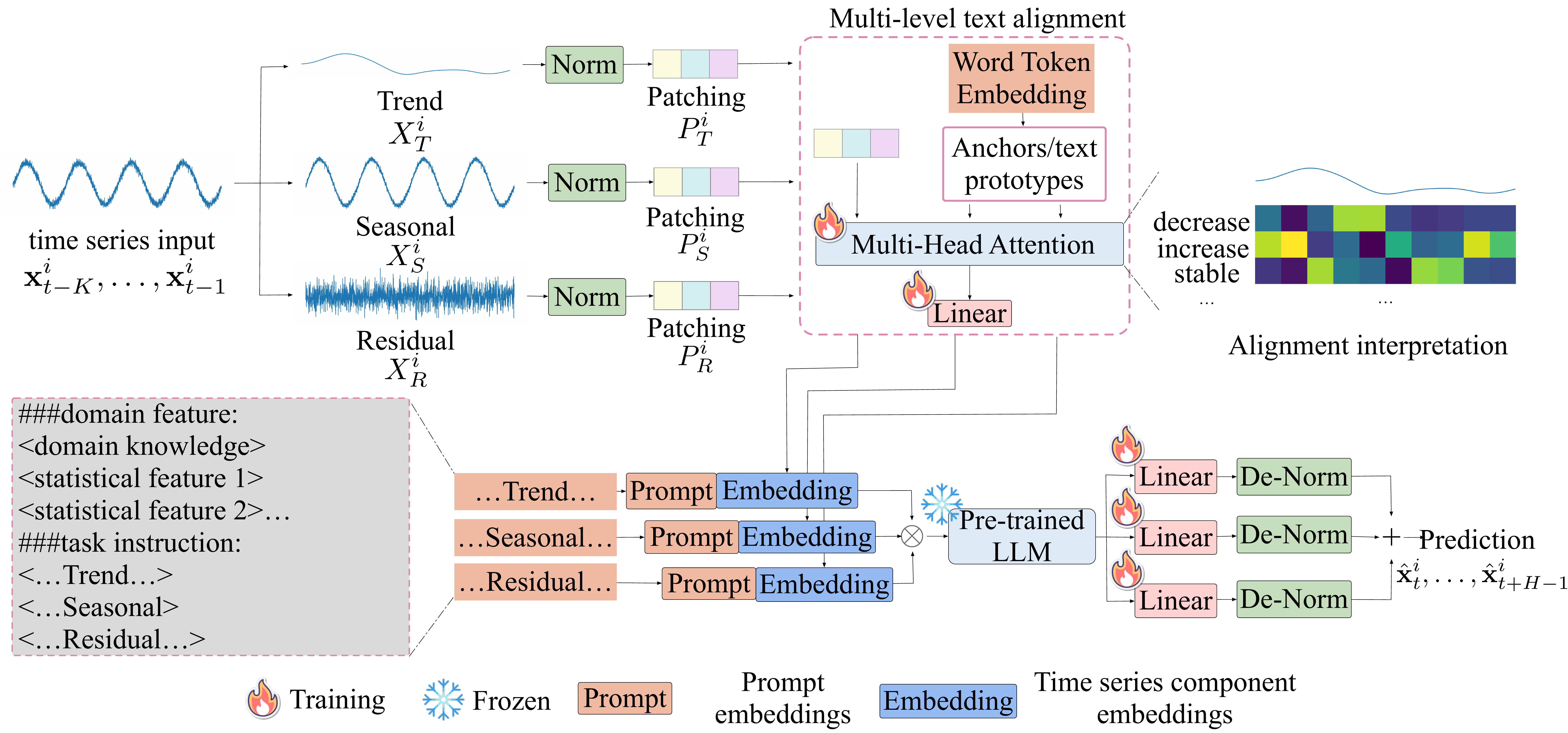}
\caption{The architecture of the proposed multi-level aligned embeddings begins with the decomposition of the input time series into three components: trend, seasonal, and residual. These components, tokenized and embedded, are reprogrammed using anchors and condensed text prototypes to align time series data with word tokens. Component-specific prefixed prompts are added to guide the transformation of input patches. Outputs from the LLM are projected, de-normalized, and summed to generate the final prediction.}
\label{overview}
\vspace{-1em}
\end{figure*}

\section{Methodology}
Our approach enhances the interpretability of LLMs for time series data using multi-level aligned embeddings. As shown in Figure \ref{overview}, the framework consists of four core modules: (1) time series decomposition, (2) multi-level text alignment, (3) component-specific prompts, and (4) output projection. A multivariate time series is first partitioned into $N$ univariate series, each processed independently as $\mathbf{X}^{(i)} \in \mathbb{R}^{1\times L}$. The series undergoes decomposition, normalization, patching, and embedding before alignment with anchor points and text prototypes. To enhance reasoning, we introduce component-specific prompts, enabling the model to generate meaningful representations, which are projected through a linear layer to produce the final forecasts, $\hat{\mathbf{x}}_t^{(i)}, \dots, \hat{\mathbf{x}}_{t+H-1}^{(i)}$. We utilize GPT-2~\cite{gpt2}, employing its first six layers as the backbone for forecasting without fine-tuning.

\subsection{Problem statement}
Given the previous $L$ observed values, the multivariate time series forecasting is predicts the values for the next $H$ timestamps. It can be formally as:
\begin{equation}
    \hat{\mathbf{x}}_t^{(i)}, \dots, \hat{\mathbf{x}}_{t+H-1}^{(i)} = F\left(\mathbf{x}_{t-L}^{(i)}, \dots, \mathbf{x}_{t-1}^{(i)}; \mathbf{V}^{(i)} \right),
\end{equation}
where $\hat{\mathbf{x}}_t^{(i)}, \dots, \hat{\mathbf{x}}_{t+H-1}^{(i)}$ is the vector of $H$-step prediction from timestamp $t$ of channel $i$ corresponding to the $i$-th feature. Given the historical values $\mathbf{x}_{t-L}^{(i)}, \dots, \\ \mathbf{x}_{t-1}^{(i)}$, a large language model $F$ uses prompt  $\mathbf{V}^{(i)}$ to make these predictions. Leveraging the strong reasoning capabilities of pre-trained LLMs, we aim to align time series data with text to enable LLMs to interpret the input series and accurately forecast the $H$ future steps, with the overall objective expressed as: 
\begin{equation}
\frac{1}{H} \sum_{h=1}^{H} \| \hat{\mathbf{x}}_{t-1+h}^{(i)} - \mathbf{x}_{t-1+h}^{(i)}\Vert^{2}.
\end{equation}

\subsection{Time series input decomposition}
For time series data, decomposing complex inputs into meaningful components such as trend, seasonal, and residual elements can help optimally extract valuable information. In this paper, given the input $X \in \mathbb{R}^{N\times L}$, where $N$ is the feature size and $L$ is the length of the time series, the additive decomposition can be represented as:
\begin{equation}
    X^{(i)} = X^{(i)}_T + X^{(i)}_S + X^{(i)}_R,
\end{equation}
where $i$ refers to the feature index for multivariate time series input. The trend component $X_T \in \mathbb{R}^{N \times L}$, capturing the underlying long-term patterns in the data, is expressed as $ X_T = \frac{1}{m} \sum_{j=-k}^{k} X_{t+j}$, where $m=2k+1$ and $k$ is the averaging step size. The seasonal component $X_S \in \mathbb{R}^{N \times L}$ reflects the repeating short-term cycles and can be estimated after removing the trend. The residual component $X_R \in \mathbb{R}^{N \times L}$ represents the remainder. Additive decomposition can be performed using various methods. A common approach first extracts the long-term trend via moving averages, estimates the seasonal component by averaging the detrended series, and obtains the residual by subtraction. Another widely used method is Seasonal-Trend decomposition using Loess (STL)~\cite{cleveland1990stl}. The decomposition method in this paper is chosen based on validation results.

Following the approach outlined in~\cite{patchtst}, we patch the decomposed components of the time series. Specifically, for the $i$-th normalized trend component, we obtain the patched token $\textbf{P}^{(i)}_T \in \mathbb{R}^{L_P\times K}$, where $L_P$ represents the patch length and $K = \lfloor \frac{(L-L_P)}{s}\rfloor + 2$ denotes the number of patches, with $s$ is the stride. Similarly, we apply this patching process to the seasonal and residual components, obtaining patched tokens $\textbf{P}^{(i)}_S$ and $\textbf{P}^{(i)}_R$, respectively. These patches are then fed into the multi-level text alignment to produce aligned time series embeddings.

\subsection{Multi-level text alignment}
We reprogram patch embeddings into the LLMs' pre-training space to align time series and natural language modalities, enhancing the backbone's reasoning capabilities. A naive approach aligns time series and text embeddings via similarity estimation, placing time series tokens near relevant text descriptors (e.g., up, down, stable). However, the pre-trained word embedding space is vast and dense, and selecting appropriate text prototypes is often arbitrary, sometimes involving unrelated words or clusters~\cite{sun2023test,s2ip,timellm}, leading to poor interpretability.

In this work, we propose multi-level text alignment to enhance the interpretability of LLMs on time series forecasting. We first decompose the time series into trend, seasonal, and residual and align the trend $X_T$ with selected trend-specific anchors $\mathbb{W}_{trend}$. However, accurately defining anchors for the seasonal and residual components is challenging. To address this, we reprogram seasonal $X_S$ and residual $X_R$ using pre-trained word embeddings $\textbf{E}\in\mathbb{R}^{V\times D}$ in the backbone, where $V$ is the vocabulary size, $D$ is the hidden dimension of the pre-trained LLM. Directly leveraging $\textbf{E}$ will result in large and potentially dense reprogramming space. We adapt linearly probing $\textbf{E}$. The text prototypes of seasonal and residual denoted as $\textbf{E}^{'}_{seasonal}\in \mathbb{R}^{V^{'}_{seasonal}\times D}$ and $\textbf{E}^{'}_{residual}\in \mathbb{R}^{V^{'}_{residual}\times D}$, where $V^{'}_{seasonal} < V^{'}_{residual}\ll V$ because the residual is more inconsistent and variable compared to the seasonal. 

As illustrated in the top-right of Figure \ref{overview}, our multi-level text alignment aims to give a connection between anchors and trend patches. The selected anchors are sparse. We reprogram seasonal and residual with text prototypes connecting time series patches with a more dense reprogramming space. To realize this, we employ a multi-head cross-attention layer for each component. Specifically, for $i$-th input feature, we define query matrices $\mathbf{Q}_{T}^{(i)} = \textbf{P}_{T}^{(i)}\textbf{W}^{Q(i)}_{T}$, key matrices $\mathbf{K}_{T}^{(i)} = \mathbb{W}_{trend}\textbf{W}^{K(i)}_{T}$, value matrices $\mathbf{V}_{T}^{(i)} = \mathbb{W}_{trend}\textbf{W}^{V(i)}_{T}$ for trend; 
query matrices $\mathbf{Q}_{S}^{(i)} = \textbf{P}_{S}\textbf{W}^{Q(i)}_{S}$, key matrices $\mathbf{K}_{S}^{(i)} = \textbf{E}^{'}_{seasonal}\textbf{W}^{K(i)}_{S}$, value matrices $\mathbf{V}_{S}^{(i)} = \textbf{E}^{'}_{seasonal}\textbf{W}^{V(i)}_{S}$ for seasonal; query matrices $\mathbf{Q}_{R}^{(i)} = \textbf{P}_{R}\textbf{W}^{Q(i)}_{R}$, key matrices $\mathbf{K}_{R}^{(i)} = \textbf{E}^{'}_{residual}\textbf{W}^{K(i)}_{R}$, value matrices $\mathbf{V}_{R}^{(i)} = \textbf{E}^{'}_{residual}\textbf{W}^{V(i)}_{R}$ for residual. Through multi-head attention, we reprogram each time series component.
\begin{equation}
\mathbf{Z}_{T}^{(i)} = \text{ATTENTION}(\mathbf{Q}_{T}^{(i)}, \mathbf{K}_{T}^{(i)}, \mathbf{V}_{T}^{(i)}) = 
\text{SOFTMAX}\left(\frac{\mathbf{Q}_{T}^{(i)} \left(\mathbf{K}_{T}^{(i)}\right)^\top}{\sqrt{d_k}}\right) \mathbf{V}_{T}^{(i)}
\end{equation}
where $d_k$ is the dimension of each head in the multi-head attention module. 

\subsection{Component-specific prompts \& output projection}
Prompting techniques effectively encode task-specific knowledge, improving model accuracy and coherence. However, translating time series into natural language is challenging, complicating instruction-following datasets and real-time prompting~\cite{xue2022prompt}. Recent advances show that prompts can enrich input context and guide reprogrammed time series patches~\cite{timellm}. To leverage semantic information, we propose a component-specific prefix prompting strategy, incorporating dataset context, statistical features, and task instructions for trend, seasonal, and residual components. For example, the instruction 'forecast the next 96 steps given the previous 512 steps [trend, seasonal, residual]' serves as a template.

After processing these prompts through the backbone LLM, we retain the component embeddings and apply a linear projection. By denormalizing and summing these representations, we derive the final forecasts $\hat{\mathbf{x}}_t^{(i)}, \dots, \hat{\mathbf{x}}_{t+H-1}^{(i)}$.

\section{Experiments}
In our experiments, the proposed method outperforms state-of-the-art forecasting approaches across various benchmarks, including long-term, short-term, and few-shot forecasting. For a fair comparison, we follow the configurations outlined in~\cite{wu2022timesnet} across all baselines, utilizing a unified evaluation pipeline\footnote{https://github.com/thuml/Time-Series-Library}. 

\textbf{Baselines.} We compare with the SOTA time series models and cite their performance from~\cite{gpt4ts, chang2023llm4ts} if applicable. The baselines include Transformer-based models like PatchTST~\cite{patchtst}, non-transformer-based techniques, i.e., DLinear~\cite{DLinear}, and four LLM-based approaches:  TimeLLM~\cite{timellm}, LLM4TS~\cite{chang2023llm4ts}, GPT4TS~\cite{gpt4ts}, and LLMTime~\cite{gruver2024LLMTime}. Aligned with the GPT4TS configuration~\cite{gpt4ts}, we utilize only the first 6 layers of the 12-layer GPT-2 base~\cite{gpt2} as the backbone for our model and TimeLLM.

\textbf{Long-term Forecasting.} For long-term forecasting, we evaluate on ETTh1, ETTh2, ETTm1, ETTm2, Weather, Electricity(ECL), and Traffic, which have been widely adopted as benchmarking datasets for long-term forecasting works~\cite{wu2022timesnet}. The input time series length $L$ is set as 512, and we evaluate across four prediction horizons: $H \in \{96, 192, 336, 720\}$. The evaluation metrics include mean square error (MSE) and mean absolute error (MAE).

\noindent Table \ref{full long term forecasting full results} presents the performance of various time series forecasting models on MSE and MAE metrics across different prediction horizons on multiple benchmarks. Our proposed model consistently outperforms existing baselines, demonstrating superior performance on average across most datasets and prediction lengths. This highlights the applicability of multi-level text alignment. Notably, our comparison with TimeLLM—a recent work leveraging text prototype reprogramming to align time series with text tokens—is significant. Specifically, our model achieves substantial improvements on the Weather and ETTm1 datasets, exceeding the best-performing model, LLM4TS, by \textbf{23.3\%} and \textbf{26.8\%}, respectively, in terms of MSE. Additionally, it records the lowest error rates across numerous dataset-prediction length configurations. These results suggest that integrating LLMs with multi-level text alignment can significantly enhance the accuracy of long-term time series forecasting. We also conduct a few-shot forecasting. More detailed experimental results for long-term forecasting and few-shot forecasting are in the appendix and full paper~\cite{FullPaper}.
\begin{table}[t]
\centering
\caption{Long-term forecasting results. All results are averaged from four different forecasting horizons:  $H \in \{96, 192, 336, 720\}$ horizons. Lower values indicate better performance. \textbf{bold}: the best, \underline{underline}: second best.}
\resizebox{\columnwidth}{!}{
\begin{tabular}{c|cc|cc|cc|cc|cc|cc}
\toprule
Methods & \multicolumn{2}{c|}{Ours} & \multicolumn{2}{c|}{Time-LLM }& \multicolumn{2}{c|}{LLM4TS} & \multicolumn{2}{c|}{GPT4TS} & \multicolumn{2}{c|}{DLinear} & \multicolumn{2}{c}{PatchTST} \\
\midrule
Metric & MSE & MAE & MSE & MAE & MSE & MAE & MSE & MAE & MSE & MAE & MSE & MAE \\
\midrule
ETTh1 & 0.424 & 0.448 & 0.420 & 0.438 & \textbf{0.404} & \textbf{0.418} & 0.428 & \underline{0.426} & 0.422 & 0.437 & \underline{0.413} & 0.435\\
\midrule
ETTh2 & 0.378 & 0.408 & 0.364 & 0.403 & \underline{0.333} & \underline{0.383} & 0.355 & 0.394 & 0.431 & 0.446 & \textbf{0.330} & \textbf{0.379}\\
\midrule
ETTm1 & \textbf{0.251} & \textbf{0.325} & 0.351 & 0.383 & \underline{0.343} & \underline{0.378} & 0.352 & 0.383 & 0.357 & 0.378 & 0.351 & 0.380\\
\midrule
ETTm2 & \textbf{0.214} & \textbf{0.289} & 0.270 & 0.328 & \underline{0.251} & \underline{0.313} & 0.267 & 0.326 & 0.267 & 0.333 & 0.255 & 0.315\\
\midrule
Weather & \textbf{0.171} & \textbf{0.226} & 0.229 & 0.267 & \underline{0.223} & \underline{0.260} & 0.237 & 0.271 & 0.248 & 0.300 & 0.225 & 0.264\\
\midrule
ECL & \textbf{0.159} & 0.262 & 0.164 & 0.261 & \underline{0.159} & \underline{0.253}  & 0.167 & 0.263 & 0.166 & 0.263 & 0.161 & \textbf{0.252}\\
\midrule
Traffic & \textbf{0.346} & \underline{0.265} & 0.408 & 0.290 & 0.401 & 0.273  & 0.414 & 0.294 & 0.433 & 0.295 & \underline{0.390} & \textbf{0.263}\\
\bottomrule
\end{tabular}
}
\label{full long term forecasting full results}
\end{table}

\textbf{Zero-shot forecasting.} Beyond few-shot learning, LLMs also show promise as effective zero-shot learners. In this section, we evaluate the zero-shot learning capabilities of the multi-level text-aligned LLM. Specifically, we assess how well the model performs on one dataset after being optimized on another. Similar to the few-shot learning setup, we use the long-term forecasting protocol and evaluate various cross-domain scenarios utilizing the ETT datasets.
\begin{table}[t]
\centering
\caption{Zero-shot learning results on ETT datasets. All results are averaged from four different forecasting horizons: $H \in \{96, 192, 336, 720\}$.}
\resizebox{\columnwidth}{!}{
\begin{tabular}{l|cc|cc|cc|cc|cc|cc}
\toprule
{Methods} & \multicolumn{2}{c|}{Ours} & \multicolumn{2}{c|}{Time-LLM}& \multicolumn{2}{c|}{GPT4TS} & \multicolumn{2}{c|}{LLMTime} & \multicolumn{2}{c|}{PatchTST} & \multicolumn{2}{c}{DLinear}  \\
\midrule
 {Datasets} & MSE & MAE & MSE & MAE & MSE & MAE & MSE & MAE & MSE & MAE & MSE & MAE \\
\midrule
{ETTh1 $\rightarrow$ ETTh2} 
 & \textbf{0.346} & \textbf{0.396} & \underline{0.354} & \underline{0.400} & 0.406 & 0.422 & 0.992 & 0.708 & 0.380 & 0.405 & 0.493 & 0.488\\
\midrule
{ETTh1 $\rightarrow$ ETTm2}  
 & \textbf{0.294} & \textbf{0.357} & \underline{0.310} & 0.363 & 0.325 & 0.363 & 1.867 & 0.869 & 0.314 & \underline{0.360} & 0.415 & 0.452\\
\midrule
{ETTh2 $\rightarrow$ ETTm2}  
 & \textbf{0.276} & \textbf{0.345} & \underline{0.303} & \underline{0.356} & 0.335 & 0.370 & 1.867 & 0.869 & 0.325 & 0.365 & 0.328 & 0.386\\
\midrule
{ETTm1 $\rightarrow$ ETTm2}  
 & \textbf{0.217} & \textbf{0.284} & \underline{0.275} & \underline{0.325} & 0.313 & 0.348 & 1.867 & 0.869 & 0.296 & 0.334 & 0.335 & 0.389\\
\midrule
{ETTm2 $\rightarrow$ ETTm1}  
 & \underline{0.562} & \underline{0.478} & \textbf{0.501} & \textbf{0.453} & 0.769 & 0.567  & 1.933 & 0.984 & 0.568 & 0.492 & 0.649 & 0.537\\
\bottomrule
\end{tabular}
}
\label{brief zero shot learning forecasting}
\end{table}

\noindent The brief results are presented in Table \ref{brief zero shot learning forecasting}. Our model demonstrates performance that is comparable to or surpasses other baselines. In data-scarce scenarios, our model significantly outperforms other LLM-based models, consistently providing better forecasts. Both our model and TimeLLM~\cite{timellm} outperform traditional baselines, likely due to cross-modality alignment, which more effectively activates LLMs’ knowledge transfer and reasoning capabilities for time series tasks. Additionally, our multi-level aligned embeddings better align language cues with temporal components of time series, enabling superior zero-shot forecasting performance compared to TimeLLM. More detailed experimental results for zero-shot forecasting are in the appendix and full paper~\cite{FullPaper}.

\begin{wraptable}{r}{0.55\columnwidth} 
    \centering
    \resizebox{0.55\columnwidth}{!}{ 
    \begin{tabular}{l|cc|cc}
        \toprule
        \multirow{2}{*}{Variant} & \multicolumn{2}{c|}{Long-term} & \multicolumn{2}{c}{Few-shot} \\
        \cline{2-5}
         \rule{0pt}{12pt} & 96 & 192  & 96 & 192 \\
        \midrule
        \textbf{Default} GPT-2 (6) & 0.117 & 0.198 & 0.360 & 0.429 \\
        \midrule
        \textbf{A.1} w/o alignment & 0.262 & 0.347 &  0.571 & 0.583 \\
        \midrule
        \textbf{B.1} only trend align & 0.184 & 0.283 & 0.476 & 0.578 \\
        \textbf{B.2} only seasonal align & 0.127 & 0.212  & 0.367 & 0.432 \\
        \textbf{B.3} only residual align & 0.171 & 0.229  & 0.433 & 0.506 \\
        \midrule
        \textbf{C.1} noise anchors & 0.134 & 0.214 & 0.424 & 0.464 \\
        \textbf{C.2} synonymous anchors & 0.119 & 0.202 & 0.366 & 0.434 \\
        \midrule
        \textbf{D.1} w/o instruction & 0.125 & 0.205 & 0.408 & 0.461\\
        \textbf{D.2} w/o domain features & 0.118 & 0.199 & 0.371 & 0.440 \\
        \bottomrule
    \end{tabular}
    }
    \caption{Comparison of different variants for long-term and few-shot forecasting on ETTm1 dataset with horizons 96 and 192.}
    \label{ablation study}
\end{wraptable}

\textbf{Multi-level text alignment variants.} 
Table \ref{ablation study} shows that removing component alignment or prefixed prompts hinders knowledge transfer in LLM reprogramming for time series forecasting. Without alignment (\textbf{A.1}), performance drops by 75.4\% on average. Retaining only seasonal alignment (\textbf{B.2}) performs best but still increases MSE by 4.5\%, while trend-only alignment causes a 32.2\% loss. Using noise anchors (\textbf{C.1}) raises MSE by 14.5\%, whereas synonymous anchors (\textbf{C.2}) yield similar results to the default with under 2\% variation. Removing component-specific instruction (\textbf{D.1}) and domain features (\textbf{D.2}) increases MSE by 7.7\% and 1.7\%, respectively.

\begin{wrapfigure}{r}{0.6\textwidth}
    \centering
    \includegraphics[width=0.6\textwidth]{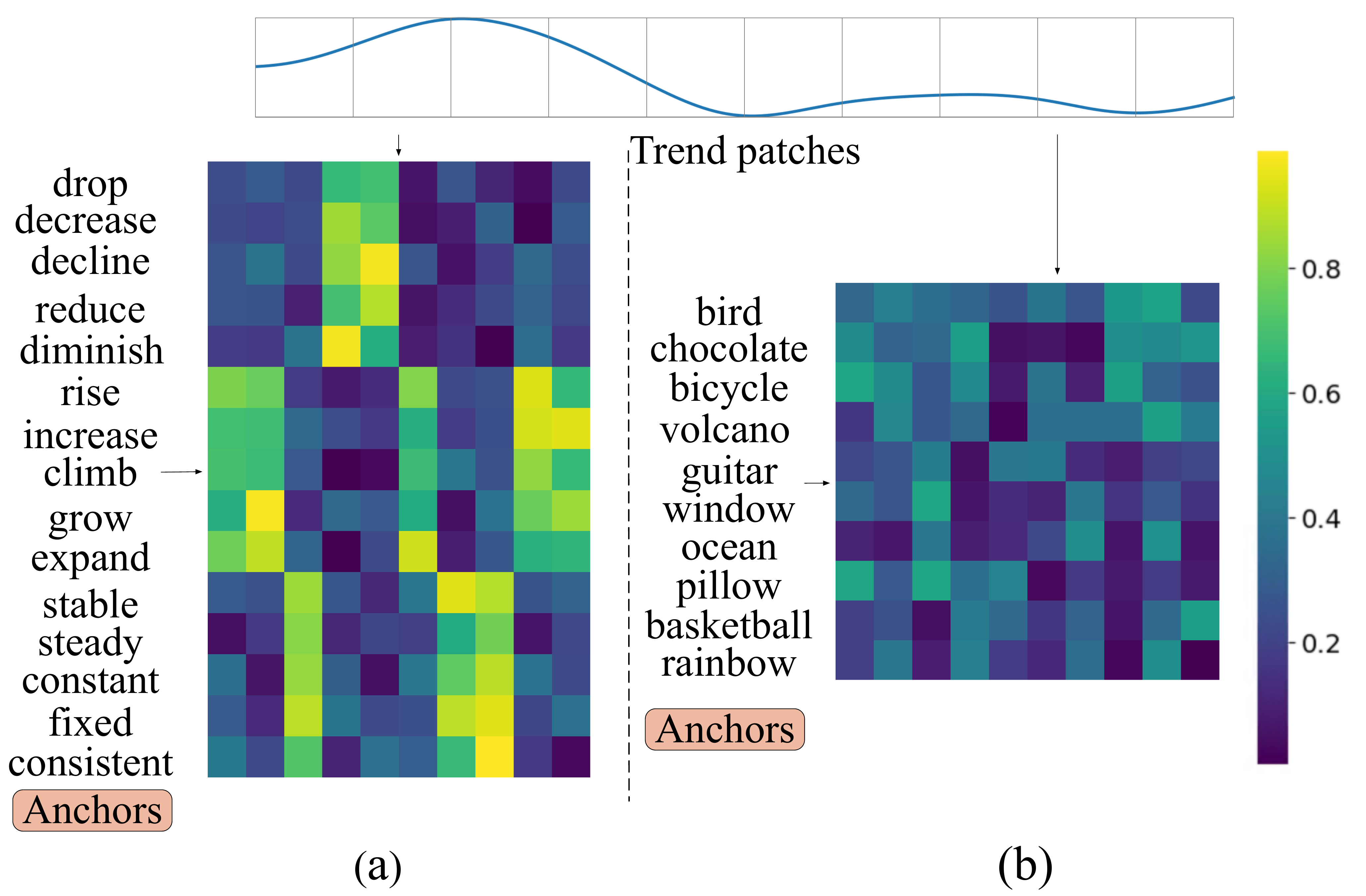}
    \caption{A showcase of visualization of multi-level alignment interpretation.}
    \label{visualization interpretation}
\end{wrapfigure}

\noindent\textbf{Multi-level text alignment interpretation.}
We conduct a case study on ETTm1 using non-overlapping patching, where the patch stride equals the patch length. Figure \ref{visualization interpretation} shows the selected anchors, including ``increase, decrease, upward, downward, linear, exponential, drift, stable, volatile, stationary, persistent, and rapid''. The attention map highlights optimized attention scores between trend patches and aligned anchors, serving as textual shapelets for time series tokens.

The attention map illustrates the optimized attention scores between input trend patches and aligned anchors. These matched anchors serve as textual shapelets for the time series tokens. Specifically, subplot \textbf{(a)} displays the optimized attention scores for synonymous anchors, which are consistent. The highlighted anchors, ``rise'',  ``increase'', ``climb'', ``grow'', and ``expand'', are associated with upward trend patches. In contrast, subplot \textbf{(b)} shows no highlighted anchor when all trend patches are aligned with noise words unrelated to time series trends. This case study demonstrates that aligned anchors effectively summarize the textual shapelets of the input trend patches.

Although multi-level alignment in both seasonal and residual components can provide visual interpretations, visualizing these alignments is challenging. Since the input patches are aligned with text prototypes learned from a large and dense pre-trained word embedding space, more tools are needed to present a better visualization across two optimized layers. Moreover, the trend component is the most interpretable and semantically clear of the three components, in contrast to the noisy residual and the seasonal component, which lacks textual semantics. Our model efficiencies in terms of parameters, memory, and speed are comparable to TimeLLM~\cite{timellm} with only two additional lightweight aligned embedding layers for integrating trend and seasonal components.

\section{Conclusion and future work}
We propose a multi-level text alignment framework with pre-trained language models for time series forecasting. Our multi-level aligned embeddings enhance the LLM's interpretability and forecasting performance by aligning time series components with anchors and text prototypes. Our results demonstrate that time series tokens aligned with anchors provide a clearer and more intuitive interpretation of similar time series trends. Future research should focus on optimizing the alignment module for selected anchors and time series tokens, and work toward developing multimodal models capable of joint reasoning across time series, natural language, and other modalities. 
\newpage
\section{Acknowledgments}
This work was supported in part by the NSF under Grant No. 1943486, No. 2246757, No. 2315612, and 1946231.
\bibliographystyle{DASFAA}
\bibliography{DASFAA}

\appendix

\end{document}